\pdfoutput=1

\documentclass[11pt]{article}

\usepackage[final]{acl}

\usepackage{times}
\usepackage{latexsym}
\usepackage{xspace}                 
\usepackage{float}

\usepackage[T1]{fontenc}

\usepackage[utf8]{inputenc}

\usepackage{microtype}

\usepackage{inconsolata}

\usepackage{graphicx}

\usepackage{soul}
\sethlcolor{lblue}
\usepackage{makecell}
\usepackage{booktabs}
\usepackage{dsfont}
\usepackage{fdsymbol}
\usepackage{multirow}
\usepackage{colortbl}

\newcommand{\hlc}[2][yellow]{{%
    \colorlet{foo}{#1}%
    \sethlcolor{foo}\hl{#2}}%
}
\colorlet{lblue}{blue!15}
\colorlet{lviolet}{violet!15}
\colorlet{lgreen}{green!35}

\title{CAVE: Controllable Authorship Verification Explanations}

\author{Sahana Ramnath$^{\varheartsuit}$ Kartik Pandey$^{\varheartsuit}$ Elizabeth Boschee$^{\clubsuit}$ Xiang Ren$^{\varheartsuit}$ \\
$^{\varheartsuit}$Department of Computer Science, University of Southern California\\$^{\clubsuit}$Information Sciences Institute, University of Southern California\\
\texttt{sramnath@usc.edu} \\
}

\newcommand{\method}{\textsc{Cave}\xspace}
\newcommand{\pcave}{\textsc{Prompt-CAVE}\xspace}
\newcommand{\av}{\textsc{Av}\xspace}
\newcommand{\promptav}{\textsc{PromptAV}\xspace}
\newcommand{\lip}{\textsc{Lip}\xspace}
\newcommand{\chot}{\textsc{CoT}\xspace}

\newcommand{\gptft}{\textsc{GPT-4-Turbo}\xspace}

\newcommand{\llamath}{\textsc{LLaMa-3-8B}\xspace}
\newcommand{\llamai}{\textsc{LLaMa-3-8B-Instruct}\xspace}

\newcommand{\imdb}{\textsc{IMDb62}\xspace}
\newcommand{\blog}{\textsc{Blog-Auth}\xspace}
\newcommand{\ff}{\textsc{FanFiction}\xspace}

\newcommand{\acc}{\textsc{Accuracy}\xspace}
\newcommand{\cons}{\textsc{Consistency}\xspace}
\newcommand{\clr}{\textsc{Cons-R-L}\xspace}
\newcommand{\consone}{\textsc{Cs-1}\xspace}
\newcommand{\constwo}{\textsc{Cs-2}\xspace}
\newcommand{\accs}{\textsc{Acc.}\xspace}
\newcommand{\conss}{\textsc{Cons.}\xspace}

\begin{document}
\maketitle

\begin{abstract}

Authorship Verification (\textbf{\av}) (do two documents have the same author?) is essential in many real-life applications. \av is often used in  privacy-sensitive domains that require an offline proprietary model that is deployed on premises, making publicly served online models (APIs) a suboptimal choice. Current offline AV models however have lower downstream utility due to limited accuracy (eg: traditional stylometry \av systems) and lack of accessible post-hoc explanations.
In this work, we address the above challenges by developing a \textit{trained, offline} 
model \textbf{\method} (\textbf{C}ontrollable \textbf{A}uthorship \textbf{V}erification \textbf{E}xplanations): \method generates free-text \av explanations that are controlled to be (1) \textit{accessible} (uniform structure that can be decomposed into sub-explanations grounded to relevant linguistic features), and (2) easily verified for explanation-label \textit{consistency}.\
We generate silver-standard training data grounded to the desirable linguistic features by a prompt-based method \textbf{\pcave}. We then filter the data based on rationale-label consistency using a novel metric \textbf{\clr}. Finally, we fine-tune a small, offline model (\llamath) with this data to create our model \textbf{\method}. 
Results on three difficult \av datasets show that \method generates high quality explanations (as measured by automatic and human evaluation) as well as competitive task accuracy\footnote{\href{https://github.com/INK-USC/Controllable-AV-Explanations}{github.com/INK-USC/Controllable-AV-Explanations}}.

\end{abstract}

\section{Introduction} \label{sec:intro}

Authorship Verification (\av) \cite{koppel2014determining} is the NLP task of determining if two input documents were written by the same author. \av is used for tasks such as plagiarism detection, forensic analysis (often in support of law enforcement), analysis of the spread of misinformation.   
Given the sensitive nature of this task, it is imperative to develop \av methods that are explainable, accessible and secure.  Explainability ensures that the system's decisions can be understood and trusted by the user. Accessibility ensures easier downstream usability; for example, a well-structured free-text explanation with clearly defined topic points is more easily understood than unstructured explanations such as long lists of low-level features. Finally, security is necessary to safely use the \av system for proprietary data; a locally-hosted model on the user's server is more secure than an API call to an online model, which could potentially store the user's sensitive data. \av is also an extremely difficult task for humans, in contrast to other NLP tasks such as machine translation; this further necessitates the need for automatic, high-quality \av systems.

\begin{figure}
    \centering
    \includegraphics[width=0.98\linewidth]{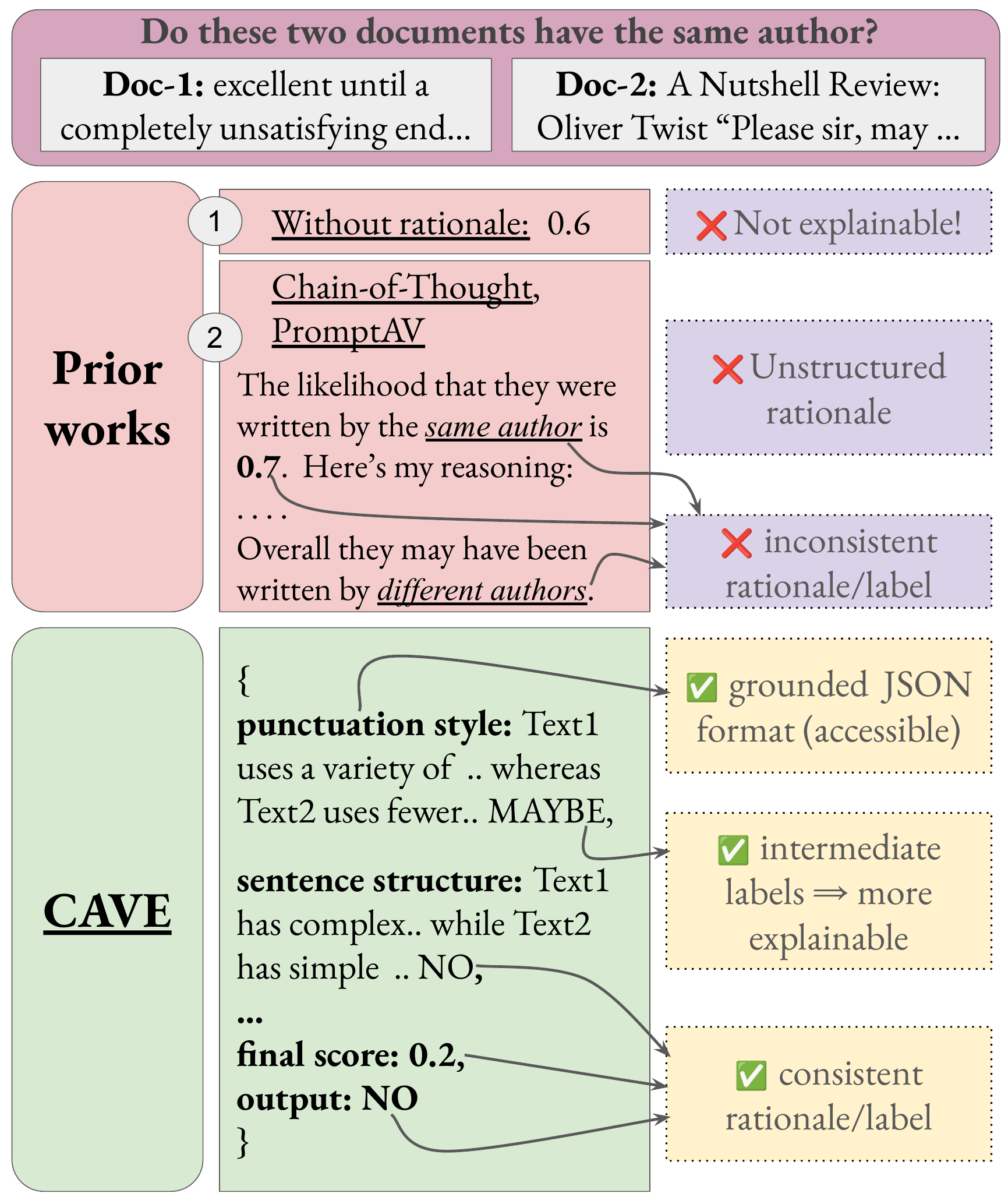}
    \caption{\textbf{\method} generates uniformly structured free-text explanations grounded in relevant linguistic features, that can be automatically verified for consistency.}
    \label{fig:motiv}
\end{figure}

\begin{table*}[h!]
    \centering
    \small

    \begin{tabular}{p{0.98\textwidth}}
    \toprule
    \{ \\
    \hl{punctuation style}: Both texts use a variety of punctuation, including commas, periods, and quotation marks, but Text 1 uses more diverse punctuation such as parentheses and hyphens. \colorbox{yellow}{\underline{MAYBE}} \\
    \hl{special characters style, capitalization style}: Text 1 uses continuous capitalization for emphasis (e.g., `WONDERFUL', `THRILLED'), which is not observed in Text 2. \colorbox{yellow}{\underline{NO}}, \\
    \hl{acronyms and abbreviations}: Neither text makes significant use of acronyms or unusual abbreviations. \colorbox{yellow}{\underline{YES}}, \\
    \hl{writing style}: Text 1 has a more personal, reflective style, sharing personal opinions and feelings about the movie. Text 2 provides a more detached, narrative-style review without personal input. \colorbox{yellow}{\underline{NO}}, \\
    \hl{expressions and idioms}: Both texts avoid colloquial expressions and idioms, opting for a more formal format. \colorbox{yellow}{\underline{YES}}, \\
    \hl{tone and mood}: Text 1 has a more varied tone, from enthusiasm to disappointment, while Text 2 maintains a consistent, somewhat formal and analytical tone. \colorbox{yellow}{\underline{NO}}, \\
    \hl{sentence structure}: Text 1 features a mix of short and long sentences with more complex structures, while Text 2 tends to use more uniformly structured, intermediate-length sentences. \colorbox{yellow}{\underline{MAYBE}}, \\
    \hl{any other relevant aspect}: The approach to movie critique is different; Text 1 is more about the impact on the viewer, while Text 2 focuses on plot summary and cinematic elements. \colorbox{yellow}{\underline{NO}}, \\
    \hl{final score}: \colorbox{yellow}{\underline{0.375}}, \\
    \hl{output}: \colorbox{yellow}{\underline{NO}} \\
    \} \\
    \bottomrule
    \end{tabular}
    \caption{Output structure of \method, \pcave: The text (JSON keys) highlighted in blue depict the linguistic features used in the analysis, as well as the keys for final confidence score and predicted task output. The text underlined and highlighted in yellow depict the intermediate labels with respect to each linguistic feature, the confidence score value, and the predicted task label. We use the predicted label to measure task accuracy and all of them to measure rationale-label consistency.}
    \label{tab:promptandop}    
\end{table*}

Early works performed \av by comparing hand-crafted features such as n-grams \cite{van2013basic}, POS tags \cite{moreau2013style}, LIWC features \cite{uchendu2020authorship}; while these are explainable, they are hard to scale. Subsequent works employed neural architectures to improve scalability and accuracy, such as the usage of Siamese networks \cite{araujo2020siamese,najafi2022text,boenninghoff2019explainable} to assign authorship via distance between document embeddings. These methods, however, have little or no post-hoc explainability. This poses a huge problem: one cannot, for instance, make legal assertions about the authorship of a set of questionable documents solely on the basis of ``a system said so'' - decision makers (judges, university officials, intelligence analysts, etc.) need to know \textit{why}.
Recent years have seen the surge of large language models and their self-rationalization capabilities: \citet{promptav,lip} generate free-text rationales\footnote{We use terms explanation and rationale interchangeably.} for \av from models such as GPT-4 \cite{gengpt} in a zero/few-shot manner. These methods however face issues such as unreliability of online models (varying results/details across runs, deprecation of APIs), 
inconsistency of rationales with predicted label (\ref{app:cot-promptav}), and high expenses.





In this work, we propose our model \textbf{\method} (\textbf{C}ontrollable \textbf{A}uthorship \textbf{V}erification \textbf{E}xplanations), an \textit{in-house} model that generates authorship predictions along with high-quality, \textit{free-text rationales} that explain the former post-hoc. The rationales are structured as a set of sub-explanations (refer Table \ref{tab:promptandop}) grounded to \av-relevant linguistic features \cite{promptav,boenninghoff2019explainable}; these sub-explanations have corresponding (intermediate) labels that provide further structure and also serve as a means to verify overall consistency with \method's authorship prediction. 

Since there is no human-written \av explanation data that can be used to train our in-house model, we adopt the model distillation strategy: 
\cite{west-etal-2022-symbolic,li-etal-2023-symbolic,li2024explanations} we engineer \textbf{\pcave} to generate silver-standard train data in the format described above by prompting a large, oracle model (\gptft \citet{gpt4turbo}), and use that to supervised-finetune a smaller language model (\llamath \citet{llama3}). 
We also introduce our automatic metric \textbf{\clr} that measures the consistency between the rationale and the model's final label; \clr is used to filter the silver train data before finetuning \cite{ramnath2023tailoring,li-etal-2023-symbolic}, and also assess \method during inference. 
\begin{figure*}[h]
    \centering
    \includegraphics[width=0.95\textwidth]{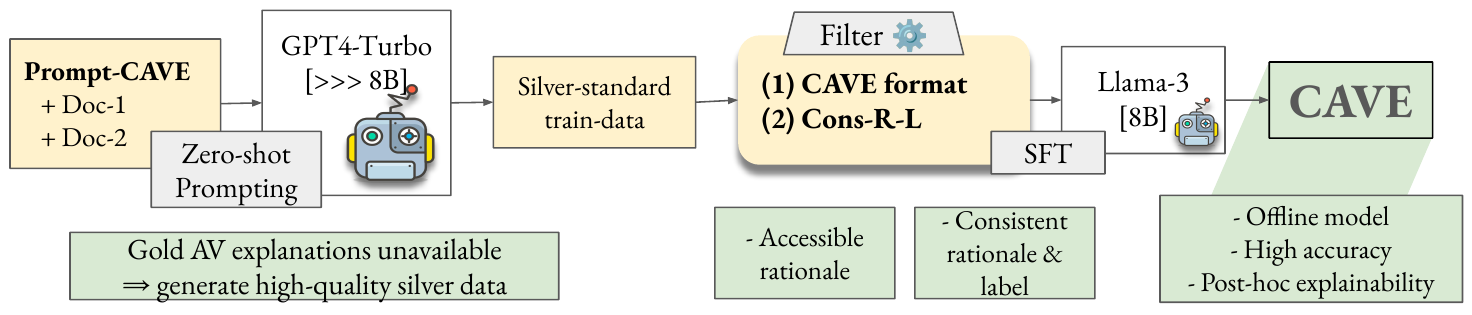}
    \caption{Pipeline to train \method: We obtain silver train data from \gptft using \pcave, filter it according to \clr and our output format. We then supervised-finetune a \llamath with the filtered data. }
    \label{fig:method}
\end{figure*}
Our experiments with three difficult \av datasets show that \method leads to competitive task performances as well as high quality of rationales.


\section{Our Method} \label{sec:method}

To enable and improve downstream utility of authorship verification systems, we present \textbf{\method}, our trained, \textit{offline} model that generates accessible free-text rationales (well-structured, and uniform structure across all datapoints). We further propose a novel metric \textbf{\clr} that leverages the structure of \method's rationales to measure consistency between a rationale and its task (\av) label. Figure \ref{fig:method} demonstrated the visual pipeline to train \method. 

\textbf{Silver-standard data.} There exists no gold-standard, human-written datasets for \av \textit{explanations}; hence, the first step to develop an \av model that can generate explanations is to create \textit{silver-standard} data for it. We propose our zero-shot prompt, \textbf{\pcave}, that can be used with LLMs such as \gptft to generate silver training data. We adapt and extend \promptav \cite{promptav} to engineer \pcave.  We then use this silver data to train an in-house language model \textbf{\method} to generate \av explanations.

\textbf{\pcave and \promptav.} Prior work \promptav proposes a zero/few-shot prompt to generate free-text rationales that contain linguistic features relevant for \av\footnote{adapted and compiled from \citet{boenninghoff2019explainable}}. To our best knowledge, \promptav was the first work in contemporary literature to generate natural language \av explanations (the second and latest work is \lip \cite{lip}, which uses a different compilation of linguistic features)\footnote{We discuss a contemporaneous work \cite{huinstructav} in Appendix \ref{sec:contemporaneous}}. While \promptav leads to high \av accuracies, its rationales are (1) inaccessible towards automatic evaluation\footnote{this is reinforced by their reporting only task accuracy and no rationale metrics}, and (2) potentially \textit{inconsistent} with their final label (detailed discussion and example in Appendix \ref{app:cot-promptav}), both of which makes the rationales unusable. In this work, we propose \pcave, an extension of \promptav that addresses these issues: (1) \pcave generates rationales that are uniformly structured across datapoints; it generates a sub-rationale and label for each linguistic feature of analysis, under a fixed JSON format. This ensures ease of understanding, and enables any automatic quality measurement. 
(2) The aforementioned intermediate labels, along with the final score and the final label (also prompted for by \pcave) allow for an \textit{automatic consistency evaluation} (described below), which leads to higher transparency and awareness.
Table \ref{tab:prompt} shows \pcave.

\textbf{Output structure (Table \ref{tab:promptandop}).} We define the \textit{correct} and \textit{complete} output structure of \method/\pcave to be a JSON structure where the linguistic features are keys; the model has to discuss the authorship of the texts grounded to each feature independently, via corresponding explanations and intermediate labels (YES/NO/MAYBE). The model also has to provide a score $\in [0,1]$ that indicates its overall confidence in the shared authorship of the documents, and an overall predicted task label (YES/NO).

\textbf{\clr, automatic consistency metric.} Rationales that are inconsistent with the model's predicted label are risky and unusable in downstream applications \cite{lyu2023faithful}. Prior works in self-rationalization check for consistency either by human analysis \cite{chen2023zara} or by training models \cite{hase2020leakage,chen-etal-2023-rev,wiegreffe-etal-2021-measuring} for the same. 
In our output format, the presence of intermediate labels allows the automatic verification of rationale-label consistency. We define a consistency metric \textbf{\clr} using two binary variables \consone and \constwo: $\clr \iff (\consone \land \constwo)$.

\textbf{\consone:} Is the final score (which indicates the model's confidence in shared authorship) consistent with the final label? If the final label is `YES', is the final score $\geq 0.5$? Similarly, if the final label is `NO', is the final score $\leq 0.5$?

\textbf{\constwo}: Are the intermediate labels faithful (overall) to the final label? Since the model output is structured as a JSON, we can automatically extract the intermediate labels for each feature and count them. If the final label is `YES' and count(YES) + count (MAYBE) > count(NO), then the labels are consistent. Similarly, if the final label is `NO' and count(NO) + count(MAYBE) > count(YES) then the labels are consistent\footnote{at this point, we assume all linguistic features have equal weightage towards shared authorship.}.

We define that is, a given output is consistent ($\clr=1$) only if the final score, final label and the aggregate of the intermediate labels all align with each other.

\textbf{Training \method.} We generate silver training data ($D_{train}$) from \gptft ($T$) using \pcave ($P$).
\begin{equation}
    D_{train}=\{(x_i, y_i^t)\} \sim T(x_i|P)
\end{equation}
where $x_i$ represents the input document pairs and $y_i^t$ represents the outputs (rationale + label) created from $T$.
We then filter $D_{train}$ according to rationale-label consistency (\clr) and conformity to our desired structure, to get our final train set $F_{train}$. Finally, we supervised fine-tune a smaller, offline language model ($S$) with $F_{train}$; for a decoder-only model like \llamath, we train the model only on the output tokens. 
\begin{equation}    
    E_{(x,y^t)\sim F_{train}} S(y^t|x)    
\end{equation}   

\section{Experiments and Results} \label{sec:exp}

\subsection{Datasets, Training, Inference} \label{sec:data}
We conduct experiments on three popular \av datasets: \imdb \cite{imdb62}, \blog \cite{blog-auth} and \ff (via the \href{https://pan.webis.de/clef20/pan20-web/author-identification.html}{PAN-2020 Authorship Verification}). Our datasets are diverse in domain (\imdb: movie reviews, \blog: website blogs, \ff: movie/book fanfiction); we carefully select these datasets to test the generalizability of our method. 
For \imdb we use the test set of 1k text pairs from \citet{promptav}. For \blog, we create our own test set by sampling 1k document pairs from the \href{https://paperswithcode.com/dataset/blog-authorship-corpus}{original blog authorship corpus}. For \ff, we sample 1k document pairs from the PAN-2020 test set (which had 14k pairs). Our test sets have equal split of YES/NO labels.

\textbf{Creating the train set.} For \imdb and \blog, the original datasets consist of documents and their anonymous author IDs; we sample document pairs with equal amounts of same/different authors to create trainsets. For \ff, we take the `small' train set provided by PAN-2020, and sample data from the same.
We start off with a train set of $\sim$1.8k documents pairs for each dataset\footnote{we ensure that there is no leakage between train/test datasets.} and combine them with \pcave and get silver rationale data from \gptft (temperature $0.0$ and $2$ responses per sample). Finally, we filter them according to the criteria in \textsection\ref{sec:method} and use them for training. Appendix \ref{app:silverdata} details the data statistics after each stage of filtering.  




\textbf{Training and Inference.} 
We supervised, fine-tune a 
\llamath 
with LoRA \cite{Hu2021LoRALA}. 
We report all hyperparameters in Appendix \ref{app:hparams}. We use greedy decoding during inference.

\begin{table*}[h!]
\centering
\resizebox{\textwidth}{!}{%
\begin{tabular}{ccccccccc}
\toprule
\multirow{2}{*}{\textbf{Model}} & \multirow{2}{*}{\textbf{Method}} & \textbf{Test} $\rightarrow$  & \multicolumn{2}{c}{\textbf{\imdb}} & \multicolumn{2}{c}{\textbf{\blog}} & \multicolumn{2}{c}{\textbf{\ff}} \\
\cmidrule(lr){3-3} \cmidrule(lr){4-5} \cmidrule(lr){6-7} \cmidrule(lr){8-9}
 &  & \textbf{Train} $\downarrow$ & \textbf{\accs} & \textbf{\clr} & \textbf{\accs} & \textbf{\clr} & \textbf{\accs} & \textbf{\clr} \\ \midrule
 \textsc{SVM} & - & Fine-tune & 59.7 & - & 56.2 & - & 55.7 & - \\ \midrule
  & \chot & Zero-shot & 73.6 & - & 62.0 & - & 54.2 & - \\
 \gptft & \promptav & Zero-shot & 73.0 & - & 62.2 & - & 57.0 & - \\
 {[$>>>$ 8B]} & \pcave & Zero-shot & \cellcolor{lblue}67.6 & \cellcolor{lblue}0.86 & \cellcolor{lblue}57.7 & \cellcolor{lblue}0.73 & \cellcolor{lblue}52.2 & \cellcolor{lblue}0.67 \\
 & \pcave & Few-shot & \cellcolor{lblue}71.3 & \cellcolor{lblue}0.94 & \cellcolor{lblue}64.0 & \cellcolor{lblue}0.97 & \cellcolor{lblue}61.7 & \cellcolor{lblue}0.86  \\
 \midrule
   & \chot & Zero-shot  & 62.4 & - & 60.7 & - & 57.5 & - \\
   \llamath & \promptav &  Zero-shot & 62.8 & - & 57.9 & - & 53.8 & - \\
   \textsc{-Instruct} & \pcave &  Zero-shot & \cellcolor{lblue}36.7 & \cellcolor{lblue}0.16 & \cellcolor{lblue}38.8 & \cellcolor{lblue}0.25 & \cellcolor{lblue}45.5 & \cellcolor{lblue}0.17 \\
   & \pcave & Few-shot & \cellcolor{lblue}53.8 & \cellcolor{lblue}0.98 & \cellcolor{lblue}62.3 & \cellcolor{lblue}0.99 & \cellcolor{lblue}48.1 & \cellcolor{lblue}0.93  \\ 
    \midrule
    \multirow{3}{*}{\llamath} & \chot & Fine-tune & 65.7 & - & 64.6$^*$ & - & 60.5 & - \\  
    & \promptav & Fine-tune & 76.2$^*$ & - & 61.3 & - & 58.7 & - \\ \cmidrule(lr){2-9} \morecmidrules \cmidrule(lr){2-9}
     & \textbf{\method} & \textbf{Fine-tune} & \cellcolor{lviolet}\textbf{74.1} & \cellcolor{lgreen}\textbf{0.99}$^*$ & \cellcolor{lviolet}\textbf{60.6} & \cellcolor{lgreen}\textbf{1.00}$^*$ & \cellcolor{lviolet}\textbf{62.6}$^*$ & \cellcolor{lgreen}\textbf{0.99}$^*$ \\  
\bottomrule
\end{tabular}%
}
\caption{Empirical results comparing baselines and \method. We note that fine-tuned \method is competitive with strong baselines such as zero-shot and few-shot \gptft, while having higher rationale quality (via consistency).}
\label{tab:exps_single}
\end{table*}

\subsection{Baselines} \label{sec:baseline}
We present and compare with multiple baselines:

\textbf{Traditional stylometry.} We present a weak baseline \cite{ikae2021unine} from traditional stylometry. We represent input document pairs as the concatenation of their TF-IDF or word-count \footnote{\texttt{sklearn.feature\_extraction.text TfidfVectorizer, CountVectorizer}} features, and train a Support Vector Machine \cite{cortes1995support} to classify their shared authorship status (hyperparameters in Appendix \ref{app:hparams}).

\textbf{Zero-shot, prior work.} We present three zero-shot baselines with \gptft (temperature $0.0$) and \llamai (greedy decoding). The first baseline is chain-of-thought \chot \cite{wei2022chain}, a standard self-rationalization method that does not include any \av specific instructions. The second baseline is \promptav \cite{promptav}. We use the prompts for \chot and \promptav reported in \citet{promptav} (Appendix \ref{app:cot-promptav}). Both these baselines report confidence scores (0-1, higher score means more confidence in shared authorship) in lieu of labels. We extract this confidence score by string matching for decimal numbers in the output; we measure accuracy by thresholding these scores at $0.5$. \textit{Consistency cannot be calculated for these two baselines}. The third zero-shot baseline is \pcave itself.

\textbf{Fine-tuned.} Similar to how we train \method, we generate silver \chot and \promptav data from \gptft, filter them according to task correctness and train a \llamath with them. We report final train set sizes in Appendix \ref{app:hparams}.

\textbf{Few-shot.} Finally, we present a few-shot baseline with \pcave on \gptft and \llamai. This baseline is strong, but undesirable owing to the high expenses of performing few-shot \av\footnote{Each document in our datasets has \textit{on-average} 1000 tokens; performing even a 4-shot baseline is highly expensive.}.  We have submitted the few shot prompts in supplementary material.

\subsection{Empirical Results} \label{sec:empiricalresults}

Table \ref{tab:exps_single} shows the empirical results. We report accuracy (\accs) for all models, and consistency (\clr) for \method and \pcave. For all 3 datasets, \method beats all relevant baselines in terms of rationale-label consistency. For \ff, \method beats all baselines (including few-shot \gptft by 0.9\%) in accuracy; for \imdb and \blog, \method obtains competitive accuracies (2\% and 4\% less than the highest accuracy, which were \promptav-SFT and \chot-SFT respectively).

\textbf{Why show \textit{two} zero/few-shot baselines?} In addition to  the strong  (online) baseline of \gptft, we also show zero/few-shot results using a less-strong local model, \llamai. The significant performance degradation there (for all cases except few-shot \blog) shows that it is insufficient merely to take existing approaches and run them using a local instruction-tuned model; our proposed distillation strategies are essential to get a high performance with local, offline models that face the disadvantage of having a \textit{much lower} parameter size.

\textbf{\pcave versus \promptav, \chot.} We investigate the drop in accuracy as we go from \chot/\promptav to \pcave in the zero-shot baseline \gptft. We observe that a majority of this drop is caused by instruction following errors, such as not generating in a JSON format, or generating a `MAYBE' as the final label instead of `YES' or `NO'. \gptft might still does produce the right \textit{score} in the prediction (which is the indicator for the predicted label used by \promptav and \chot), but since our evaluation requires the JSON format and a clear YES/NO label at the end, \gptft's performance decreases with \pcave. For \imdb, \blog and \ff respectively, the number of test datapoints (out of 1000) where \gptft doesn't follow the output format with zero-shot \pcave are 47, 153 and 148 respectively; the number of datapoints where it produces a label of `MAYBE' are 59, 45 and 103 respectively. However, as we move to \textit{few-shot} \pcave or \textit{fine-tuned} \method, we find that instruction-following errors are almost non-existent, and there is a huge improvement in accuracy as well as consistency!  Appendix \ref{app:other-local-models} further discusses practical advantages of \method over \gptft, as well as experiments with more local models.




\subsection{In-house human study:}  \label{sec:humanpilot}
\av + explanations is a highly complex task; annotators require a considerable amount of training and practice to be skillful in their evaluation.  Hence, for our evaluation, we opted to conduct a \textit{small but high-quality} \textbf{in-house human study}, instead of a large-scale study with platforms such as Amazon MTurk. 

\noindent We pick $50$ document pairs per dataset (total of $150$ pairs) and evaluate the rationales generated by their corresponding \method. We analyze the strengths \& weaknesses of \method's rationales, which help to understand the utility of these rationales to downstream applications \cite{joshi-etal-2023-machine} and to learn about potential areas for improvement \cite{van-der-lee-etal-2019-best}. We analyze the generated rationales (for each linguistic feature as in Table \ref{tab:promptandop}) via three properties as described below; we employ 3 distinct annotators per datapoint. We also require the annotators to comment on why a rationale is unsatisfactory, if they find it to be so with respect to any property; this helps to perform a more detailed analysis at the instance level.\\
\noindent \textbf{1. Detail-Consistency:} Are the details mentioned in the rationale consistent with the input documents, or are they hallucinated? (eg:- if the rationale mentions that both documents use parantheses, but the documents don't, then the details are \textit{hallucinated})\\
\noindent \textbf{2. Factual-Correctness}: Are the rationales factually correct? (eg:- if the rationale says that the text was informal in tone, while it was actually formal, or if it misinterprets an acronym to be author-specific slang, it is incorrect)\\
\noindent \textbf{3. Label-Consistency} \method's output format has a distinct rationale and intermediate label (YES/NO/MAYBE) for each linguistic feature. As defined in \textsection\ref{sec:method}, we use these intermediate labels to check the overall consistency of the rationale with the final label. In this human evaluation however, we check whether each individual rationale is consistent with its corresponding label. 

\begin{table}[h!]
\centering
\resizebox{\linewidth}{!}{%
\begin{tabular}{lccc}
\toprule
\textbf{ Linguistic Features} & \textbf{P1} & \textbf{P2} & \textbf{P3} \\ \midrule
punctuation style & 63 & 103 & 125 \\  \midrule
special characters \& capitalization style & 87 & 116 & 136  \\ \midrule
acronyms and abbreviations & 110 & 124 & 135 \\ \midrule
writing style & 136 & 137 & 145 \\ \midrule
expressions and idioms & 130 & 137 & 138 \\ \midrule
tone and mood & 140 & 138 & 145  \\ \midrule
sentence structure & 126 & 133 & 141  \\ \midrule
any other relevant tabs & 142 & 127 & 120  \\ \midrule \midrule \midrule
aggregate & 934 & 1015 & 1085 \\ \midrule  
aggregate as \% & 77.8\% & 84.6\% & 90.4\% \\ \midrule \midrule \bottomrule
\end{tabular}%
}
\caption{\textbf{Human Pilot:} This table shows the number of examples out of 150 where all 3 annotators agree that the rationale conforms to the property being analyzed. P1 - `Detail-Consistency', P2 - `Factual-Correctness', P3 - `Label-Consistency'. We report these numbers for all linguistic features used by \method.}
\label{tab:agghumaneval}
\end{table}

\begin{table*}[ht]
    \centering
    \small
    \begin{tabular}{p{0.47\textwidth}|p{0.47\textwidth}}
    \hline
    \textbf{Rationale / Label} & \textbf{Annotator's Comment} \\ \hline
    \hl{punctuation style}: Both texts use a variety of punctuation marks, including commas, parentheses, and quotation marks, but Text2 uses more complex structures such as dashes and colons. & Hallucination of punctuation marks when they don't exist in the original documents. \\
    \hline
     \hl{any other relevant aspect:} The subject matter of the two texts is vastly different, with Text1 focusing on technology and personal preference, while Text2 focuses on societal issues and personal identity. \hlc[yellow]{NO} & Similarity/difference in topics or shared platforms does not definitively imply same/different authorship - the label should have been \hlc[yellow]{MAYBE} for both these cases. \\ \hline
    \hl{special characters style, capitalization style}: Text1 shows a casual approach to capitalization, with instances of lowercase usage in contexts that typically require capitalization (e.g.,`relly' instead of`really'). Text2 adheres to standard capitalization rules. & Confusion between capitalization and misspelling/typo.  \\ \hline
    \hl{acronyms and abbreviations}: Neither text makes significant use of acronyms or unusual abbreviations, maintaining a formal tone in dialogue without informal abbreviations. & Lack of acronyms does not indicate a formal tone - this rationale is even more incorrect when the original text is informal in tone. \\ \hline 
    \hl{acronyms and abbreviations}: Neither text makes significant use of acronyms or abbreviations, which suggests a similarity in this aspect. However, the absence of such elements makes it difficult to draw a definitive conclusion.  \hlc[yellow]{YES} & The label should have been \hlc[yellow]{MAYBE} owing to lack of information with this feature.  \\ \hline
    \hl{special characters style, capitalization style}: Both texts use colloquial expressions ('buttload of dishes', 'nagged the darn thing out of Shawn'). These expressions contribute to the informal tone of the texts. & Hallucination and factually incorrect: Both expressions were in Text-1, none in Text-2. \\ \hline
    
    \end{tabular}
    \caption{Sample rationales from various examples, to show observed error types by \method}
    \label{tab:expl-eg}
\end{table*}

\noindent In Table \ref{tab:agghumaneval}, we show number of examples (out of 150) for which the three annotators agree that the rationale aligns with the property being analyzed. We also show aggregate statistics for each property across all linguistic features, i.e., $\frac{sum(P1)}{150*8}$ where 8 is the number of linguistic features being analyzed. 

\noindent \textbf{Aggregate Analysis.}  Annotators find that \method's rationales align with properties P1, P2, P3 \textbf{77.8\%}, \textbf{84.6\%} and \textbf{90.4\%} of the cases respectively. That is, \method hallucinates, presents incorrect details and produces inconsistent intermediate labels in 22.2\%, 15.4\% and 9.6\% of the cases respectively. 


\noindent \textbf{Spurious patterns.} Annotators noticed that documents in the \ff dataset all used double instead of single quotes for apostrophes (eg:- it''s instead of it's), leading to \method reasoning that these double quotes were indicative of shared authorship. Similarly, some documents in the \blog dataset used the term `urlLink' to denote the presence of hyperlinks (perhaps a formatting pattern used by the blog platform) - \method however, took the presence of this unique term as an indicator of shared authorship. Since annotators view many datapoints in their analysis, they are able to identify these spurious patterns; models such as \method and \gptft view these datapoints in isolation and hence do not dismiss these patterns as indicators of shared authorship.  

\noindent \textbf{Qualitative examples \& Annotator comments.} We report commonly observed rationale errors in Table \ref{tab:expl-eg}. Hallucination of details was a common error, especially for punctuation style and special characters / capitalization; for example, \method mentioning the presence of punctuation marks such as ellipses, parantheses, etc., when those did not exist in the document. One hypothesis is that LLMs are bad at these linguistic features since they deal with individual characters and not broad pieces of text like the other features. Another common error was the \method's usage of topic of texts as a reason to indicate or refute shared authorship of two documents; topic cannot be an absolute factor in deciding authorship \cite{wegmann2022same}, and hence such labels must be `MAYBE'. We also discuss random errors owing to computational limitations (such as limited context window in LMs) in Appendix \ref{app:randomerrors}.

\noindent We report the human study numbers for individual datasets in Table \ref{tab:humaneval}. We also release the \href{https://docs.google.com/spreadsheets/d/1npLyOAwdnIAWjmQbCip5w_Ajv34YU1v7uiUhp4CgmPc/edit?usp=sharing}{complete set of test-set rationales generated by \method}.


\subsection{Future steps} \label{sec:futureexps}
The first step towards handling these issues is to make downstream users aware about them, so that they can make an informed final decision with \method's rationales; the independence between linguistic features make it easy for users to process them separately and use them as needed. 

\paragraph{Ablation study with linguistic features.} A straightforward solution is to simply identify the linguistic feature that is the most uncertain/problematic in the training data, and remove it from the datapoints: we perform this experiment as an ablation study. We identified from analyzing the silver data that `punctuation style' has an intermediate label of `MAYBE' in roughly 60-70\% of the data for each dataset. We hypothesized that this feature was generally indecisive and unimportant; we removed it from the train data and fine-tuned a \llamath on the same - however, we found that the performance \textit{reduced} upon removal of this feature for all 3 datasets. This suggests that though this feature is generally indecisive, it does offer much needed information or context that the model depends upon; simply removing the feature will not improve the performance, and hence, we need more intricate methods such as weighing of linguistic features depending on the datapoint at hand.  

\paragraph{Targeted training.} As futute steps, we propose to perform targeted training that directly addresses the problems of (1) \textit{hallucinated details}, and (2) \textit{trivial reasonings from unimportant features}. For the hallucination issue, given that we have the input documents, we can potentially have a reward model that checks the entailment between the sub-explanations and the documents and verifies whether a detail is real or a hallucination; we can further perform reward-based learning with algorithms such as PPO \cite{Schulman2017ProximalPO}, Quark \cite{lu2022quark}. For the latter issue, we can have \textit{dynamic} weighting of linguistic features for each example, which highlight only the features that contribute meaningful similarities or differences between the documents. 

\section{Discussion} \label{sec:discussion}
\subsection{Robustness of \consone and \constwo} \label{sec:robustness}
\clr, the automatic metric used to measure
rationale-label consistency, is defined based on two binary variables (\textsection\ref{sec:method}): \consone which measures consistency between the final score and the final label, and \constwo which does the same between the final label and the aggregated intermediate labels. These variables were designed with common sense logic (such as, a final score $\geq 0.5$ must indicate a final label of YES). 
But how robust are they statistically? We analyze the silver train data from \gptft with relevant questions below. We use the term \textit{accepted} for datapoints that satisfied the consistency condition being analyzed (\consone or \constwo), and \textit{rejected} for those that were filtered out. We split the analysis into 5 categories: accepted datapoints with a final label of YES/NO, and rejected datapoints with a final label of YES/NO/MAYBE (when the predicted label is `MAYBE', it is always filtered out).

\textbf{\consone: Is 0.5 the best threshold?} In Table \ref{tab:cs1_scores} we present the average score obtained by accepted/rejected datapoints in the five categories. We note that for accepted data for all three datasets, \gptft assigns a final score of $0.86$ or higher when predicting the final label as `YES', and assigns a final score of  $0.33$ or lower when predicting the final label as `NO' - these indicate that accepted datapoints are well away from the threshold boundary. For rejected data, \gptft on average assigns a score surrounding $0.6$ for predicted labels of `NO' and `MAYBE' (there are no cases where the rejected data has a label of `YES', i.e., where the label was `YES' and the score $< 0.5$) - this shows that rejected datapoints were \textit{not} rejected just because of a harsh thresholding\footnote{harsh thresholding is when the final score is marginally crossing the boundary to be accepted, for example, a score of $0.51$ for a label of `NO' or $0.49$ for a label of `YES'}. 

\begin{table}[h]
\centering
\resizebox{\linewidth}{!}{%
\begin{tabular}{|l|c|c|c|c|c|}
\hline
\multirow{2}{*}{\textbf{Dataset}} & \multicolumn{2}{c|}{\textbf{Accepted}} & \multicolumn{3}{c|}{\textbf{Rejected}} \\
\cline{2-3} \cline{4-6}
 & YES & NO & YES & NO & MAYBE \\ \hline
\imdb & 0.89 & 0.33 & - & 0.63 & 0.65 \\ \hline
\blog & 0.86 & 0.18 & - & 0.62 & 0.58 \\ \hline
\ff & 0.87 & 0.30 & - & 0.63 & 0.62 \\ \hline
\end{tabular}%
}
\caption{Statistical analysis of \consone: Average `final score' generated by \gptft + \pcave in the silver train data.}
\label{tab:cs1_scores}
\end{table}

\begin{table}[h]
\centering
\resizebox{\linewidth}{!}{%
\begin{tabular}{|l|c|c|c|c|c|}
\hline
Predicted & \multicolumn{2}{c|}{\textbf{Accepted}} & \multicolumn{3}{c|}{\textbf{Rejected}} \\
\cline{2-3} \cline{4-6}
Label $\rightarrow$ & YES & NO & YES & NO & MAYBE \\ \hline
\multirow{2}{*}{Dataset $\downarrow$} & \% MAY. in & \% MAY. in & \multirow{2}{*}{\%NO} & \multirow{2}{*}{\%YES} & \multirow{2}{*}{\%MAY.} \\ 
& MAY.+YES & MAY.+NO & & & \\ \hline
\imdb & 15.7 & 38.8 & - & 51.2 & 45.7 \\ \hline
\blog & 22.1 & 24.6 & - & 50.0 & 54.3 \\ \hline
\ff & 17.0 & 32.1 & - & 51.3 & 43.1 \\ \hline
\end{tabular}%
}
\caption{Statistical analysis of \constwo: \% of intermediate labels}
\label{tab:cs2_labels}
\end{table}

\textbf{\constwo: How much do the intermediate MAYBE's contribute?} We require that the intermediate labels as an aggregate (with `MAYBE' as a buffer for both `YES' and `NO') side with the final label. But statistically, how much of this support comes from the MAYBE's, and how much comes from YES and NO? In Table \ref{tab:cs2_labels} we present relevant proportions of intermediate labels for the five categories.  For accepted data, we calculate the \% of MAYBE in support of YES and NO respectively, that is, \% of MAYBE in MAYBE + YES or MAYBE + NO. For rejected data, we calculate the \% of NO's in the intermediate labels for a predicted label of `YES' (that is, how many NO's were there in the intermediate labels to make them inconsistent with the final label?), \% YES's for a predicted label of `NO', and \% of MAYBE's for a predicted label of `MAYBE'. 
We note that for accepted data, the support given by `MAYBE' is on average $\sim$ 15-22\% for a final label of `YES', and $\sim$24-39\% for a final label of `NO' - this indicates that the main support for the final labels of YES/NO come respectively from intermediate labels of YES/NO. For rejected data where the predicted label was NO, the aggregate intermediate labels came roughly to $\sim$50\% YES, indicating that the inconsistency was not due to indecision of the model with MAYBE's. For rejected data where the predicted label was MAYBE, the aggregate intermediate labels are $\sim$43-54\% MAYBE, indicating a higher level of uncertainty in the model. 


\noindent \textbf{Measuring uncertainty via score and labels:} As per our logic, 
a final score close to $0.5$, and/or a high proportion of `MAYBE' intermediate labels indicate a higher uncertainty of the model in its final prediction. To validate this hypothesis, we calculate the correlation\footnote{\texttt{from scipy.stats import pearsonr}} between the fraction of MAYBE's in the intermediate labels, and 1 minus the absolute distance of the final score from $0.5$\footnote{a lesser distance to 0.5 implies high uncertainty, so it is subtracted from 1 to make the final value higher if there is higher uncertainty
}. Formally, 
\begin{align*}
v_1 &= \frac{\text{count}(\text{MAYBE})}{\text{num\_intermediate\_labels}} \\
v_2 &= 1 - \left| \text{final score} - 0.5 \right| \\
\text{correlation} &= \text{PearsonCorr}(v_1, v_2)
\end{align*}

We calculate this correlation for the \gptft train data; we obtain \imdb: 0.66, \blog: 0.8, \ff: 0.6, all with $p<<0.01$. The high correlation scores validate our hypothesis.

\subsection{Analyzing signal strengths across linguistic features}
In Table \ref{tab:maybestrength}, we present a detailed analysis of the intermediate label (MAYBE) signal strengths across all linguistic features, as measured from the training data of each dataset. Punctuation style sees a high proportion of MAYBE's (60-70\%) for all datasets, whereas features such as writing style, tone \& mood etc., have a very low proportion. On average across all linguistic features, the proportion of `MAYBE' is 20-25\%, reinforcing the conclusion from \textsection \ref{sec:robustness} that the majority support for the predicted label comes from decisive intermediate labels of `YES' and `NO', and not from the uncertain `MAYBE'.
\begin{table}[h]
\centering
\resizebox{\linewidth}{!}{%
\begin{tabular}{|l|c|c|c|}
\hline
\textbf{Ling. Feature} & \textbf{\imdb} & \textbf{\blog} & \textbf{\ff} \\ \hline 
punctuation style & 68.4 \% & 60.2 \% & 59.4 \% \\ \hline
special char./ & \multirow{2}{*}{21.2 \%} & \multirow{2}{*}{23.9 \%} & \multirow{2}{*}{20.7 \%} \\ 
capitalization & & & \\ \hline
acronyms \& abbr. & 26.6 \% & 44.5 \% & 29.1 \% \\ \hline
writing style & 6.5 \% & 5.9 \% & 7.1 \% \\ \hline
expressions/idioms & 24.6 \% & 14.4 \% & 23.2 \% \\ \hline
tone \& mood & 5.1 \% & 7.0 \% & 7.2 \% \\ \hline
sentence structure & 28.0 \% & 9.8 \% & 18.8 \% \\ \hline
any other & 17.5 \% & 8.6 \% & 8.8 \% \\ \hline \hline
avg. across features & 24.74 \% & 21.79 \% & 21.79 \% \\ \hline 
\end{tabular}%
}
\caption{Proportion of datapoints in the train set for each linguistic feature where the intermediate label is `MAYBE'. Punctuation style sees a high proportion of `MAYBE' labels (60-70\% across the datasets), whereas other features see much lower proportions.}
\label{tab:maybestrength}
\end{table}

\subsection{\method versus \gptft}
In Appendix \ref{app:silverdata}, we present analyses of the silver training data generated by \gptft as well as human evaluation results of the same. In addition, here we compare rationales generated by \method and \gptft on the test set in the following three aspects:
\paragraph{Does \method yield less MAYBE's than \gptft?} In terms of the final output label, since \method is fine-tuned on data that has only YES/NO labels as the final output, it always produces only YES/NO and never a MAYBE; this is not the case with \gptft since we zero/few-shot prompt it. In terms of the intermediate labels, we report the \% of MAYBE's averaged across linguistic features for zero-shot \gptft and \method's predictions on the test set in Table \ref{tab:mayberationales}. The proportions are found to be similar, with zero-shot \gptft having slightly more MAYBE's as its intermediate labels than \method.

\begin{table}[h]
    \centering
    \resizebox{0.9\linewidth}{!}{%
    \begin{tabular}{lcc}
    \hline
    \textbf{Dataset} & \textbf{\gptft} & \textbf{\method} \\ \hline
    \imdb & 29.22 \% & 23.71 \% \\ 
    \blog & 25.23 \% & 23.46 \% \\ 
    \ff & 26.64 \% & 20.34 \% \\ \hline
    \end{tabular}
    }
    \caption{Proportion of intermediate labels MAYBE's averaged across linguistic features for zero-shot \gptft and \method's predictions on the test set.}
    \label{tab:mayberationales}
\end{table}

\paragraph{Does \method yield less rejected samples with respect to \consone/\constwo than \gptft?} From Table \ref{tab:exps_single}, we see that \method has a higher \clr, which means that it yields less rejected samples than \gptft; in our individual \consone and \constwo analysis, we find that \method has higher values for both. We report values below for \method and zero-shot \gptft on the test set below in Table \ref{tab:conspercentage}.

\begin{table}[h]
    \centering
    \resizebox{0.98\linewidth}{!}{%
    \begin{tabular}{ccccc}
    \hline
    \textbf{Dataset} & \textbf{Model} & \textbf{\clr} & \textbf{\consone} & \textbf{\constwo} \\ \hline
    \multirow{2}{*}{\imdb} & \textsc{GPT-4-T} & 0.86 & 0.86 & 0.86 \\
     & \method & 0.99 & 0.99 & 1.0 \\ \hline
    \textsc{Blog-} & \textsc{GPT-4-T} & 0.73 & 0.73 & 0.73 \\ 
    \textsc{Auth} & \method & 1.0 & 1.0 & 1.0 \\ \hline
    \textsc{Fanfic-} & \textsc{GPT-4-T} & 0.67 & 0.67 & 0.67 \\ 
    \textsc{-tion} & \method & 0.99 & 0.98 & 0.99 \\ \hline
    \end{tabular}
    }
    \caption{\consone and \constwo values for \method and zero-shot \gptft on the test sets.}
    \label{tab:conspercentage}
\end{table}

\section{Conclusion}
Existing methods for authorship verification face a lack of \textit{accessible explainability}. In this work, we train an in-house language model \textbf{\method} to generates structured and consistent explanations for \av. We present strong experimental results on three difficult \av datasets; we also present a human pilot study on the quality of generated rationales. Finally, we provide detailed discussions on the robustness of our consistency metric, and explore errors made by \method due to factors such as hallucination, dataset biases, etc. to guide research in this area. As future work, we would like to explore reward metrics and reward-based learning for \av explanations.

\section*{Limitations}  \label{sec:lim}


\noindent \textbf{Bias propagation from \gptft to \method, and mitigation:} Since there is no human-written explanation data for \av, we used silver data generated by \gptft; however, this can lead to \textit{biases}\footnote{we adopt the definition of bias from \citet{jin-etal-2021-transferability,blodgett-etal-2020-language}: disparate model performance
on different subsets of data which are associated
with different demographic groups} present in \gptft to unconsciously propagate to \method. The datasets we use in this paper are not highly bias-prone; however, researchers/users who use our framework on their dataset might deal with biases that stem from their dataset (such as gender bias, racial bias, etc., for example: `Document-1's writing style is depictive of demographic-X whereas Document-2's writing style is depictive of demographic-Y'). 
For users dealing with sensitive data for \av, we recommend they use bias mitigation strategies from relevant literature such as Counterfactual Role Reversal \cite{gupta-etal-2022-mitigating} for gender bias, reducing bias in the upstream model before fine-tuning (UBM \cite{jin-etal-2021-transferability}, strategic pretraining \cite{feng-etal-2023-pretraining}), ensembling models with varying/opposing biases (partisan ensemble \cite{feng-etal-2023-pretraining}), etc.  


\textbf{Hallucination.} As we see in our human study, \method's rationales sometimes have \textit{hallucinated} details, especially for certain linguistic features. Hallucination is a common problem faced when using LMs \cite{ji2023survey,huang2023survey}; prior works \cite{zhang2023mitigating,shi2024trusting} have proposed task-specific strategies to alleviate the same. In our work, our proposed output format for \method allows for downstream users to analyze each linguistic feature independently, making it easier to \textit{identify} hallucinations. We caution users to be aware of this issue, and to always manually read and verify the rationales before using them.

\textbf{Completeness of rationale.} We would like to emphasize here that our models have not been explicitly trained for \textit{completeness}; that is, apart from any wrong information the generated explanation may have, it could also have missed some crucial similarity/dissimilarity between the documents. We discuss rationale properties such as completeness in Appendix \ref{app:more-rat}.

\textbf{Trust.} \citet{lipton2018mythos} discusses that the definition of \textit{trust} in a model is subjective: some users only trust well-understood models; other users might be inclined to trust well-performing models regardless of how (not) explainable they are.  A recent work \citet{sun2024trustllm} defines various \textit{dimensions} for trust for language models, such as truthfulness, safety, robustness, etc. Our improvements in terms of accessibility of rationales and automatic quality evaluation contribute to improved trust in certain dimensions; as future work, we would like to evaluate and improve on other dimensions of trust, including higher accuracy, more datasets, more OOD evaluation, etc.



\textbf{Practical limitations.} Lastly our method has practical limitations faced by any NLP task; our models are restricted by the size of the context window allowed by the model (both input and output text) (also keeping in mind restrictions due to computational costs, financial costs and GPU size available). This can lead to data missed by the model in its reasoning. 

\paragraph{Data.} All the datasets that we use in our work are released publicly for usage and have been duly attributed to their original authors. Note that while we perform authorship verification, all the datasets we use have been \textit{anonymized} by their respective creators. We do not have access to or use the names of individual people.

\section*{Reproducibility}
For all our experiments, we report (1) the complete hyperparameter setting and any bounds explored (Appendix \ref{app:hparams}) as well as the sizes and versions/pretrained-model links of all models used, (2) the time taken per experiment, and infrastructure used, (3) descriptions of datasets, and demonstrations used to sample rationales from \gptft. Further, we have released all our codes and our training/test data at \href{https://github.com/INK-USC/Controllable-AV-Explanations}{github.com/INK-USC/Controllable-AV-Explanations}.

\paragraph{\textsc{Llama-3} Usage and License.} Note that we were granted access to download and use \llamath from \url{https://huggingface.co/meta-llama/Meta-Llama-3-8B}. While we train a local \llamath, we do not release the trained model; we submit only the script and data used to train the model. Also note that we adhere to the intended use case for \llamath, i.e., ``Llama 3 is intended for commercial and research use in English. Instruction tuned models are intended for assistant-like chat, whereas pretrained models can be adapted for a variety of natural language generation tasks.''

\section*{Acknowledgements}
This research is supported in part by the Office of the Director of National Intelligence (ODNI), Intelligence Advanced Research Projects Activity (IARPA), via the HIATUS Program contract \#2022-22072200006. The views and conclusions contained herein are those of the authors and should not be interpreted as necessarily representing the official policies, either expressed or implied, of ODNI, IARPA, or the U.S. Government. The U.S. Government is authorized to reproduce and distribute reprints for governmental purposes notwithstanding any copyright annotation therein. We thank the human annotators for their help with the pilot study. Finally, we thank all our collaborators at USC INK Research Lab and ISI for their constructive feedback on this work.

\bibliography{custom}
\appendix

\section{\pcave} \label{app:pcave}
Table \ref{tab:prompt} shows the full \pcave. 
\begin{table*}[h!]
    \centering
    \small
    \begin{tabular}{p{0.95\textwidth}}
    \toprule
    \textbf{\pcave:} \\
    Task: On a scale of 0 to 1, with 0 indicating low confidence and 1 indicating high confidence, please provide a general assessment of the likelihood that Text 1 and Text 2 were written by the same author. Your answer should reflect a moderate level of strictness in scoring. Here are some relevant variables to this problem. \\
    1. punctuation style (e.g. hyphen, brackets, colon, comma, parenthesis, quotation mark) \\
    2. special characters style, capitalization style (e.g. Continuous capitalization, capitalizing certain words) \\
    3. acronyms and abbreviations(e.g. Usage of acronyms such as OMG, Abbreviations without punctuation marks such as Mr Rochester vs. Mr. Rochester,Unusual abbreviations such as def vs. definitely) \\
    4. writing style \\
    5. expressions and Idioms \\
    6. tone and mood \\
    7. sentence structure \\
    8. any other relevant aspect \\
    \hl{Provide the answer in a Python JSON format. Use the following keys for your dictionary: `punctuation style', `special characters style, capitalization style', `acronyms and abbreviations', `writing style', `expressions and Idioms', `tone and mood', `sentence structure', `any other relevant aspect', `final score'. Apart from the `final score', everything else must have a text value; also, the text should include a concluding YES/NO/MAYBE about whether the two texts are similar or not with respect to the key at hand. Finally, provide an `output' key in your dictionary, which says YES if the two texts are written by the same author, and NO otherwise.}\\\\
    Text1: excellent until a completely unsatisfying ending This movie really got me hooked. The plot about... \\\\
    Text2: A Nutshell Review: Oliver Twist ``Please sir, may I have some more?'' Ah, the immortal line from... \\
    \bottomrule
    \end{tabular}
    \caption{\pcave: We extend \citet{promptav} to improve controllability in terms of the structure/parsing of the output and automatic verification of the consistency between the rationale and the label. The highlighted part is our extension to \promptav.}
    \label{tab:prompt}
      
\end{table*}

\section{Related Work}
\subsection{Contemporaneous work} \label{sec:contemporaneous}
\textsc{InstructAV} \cite{huinstructav} is a contemporaneous work that also extends \promptav to generate silver training data to fine-tune a smaller language model. \textsc{InstructAV} uses a \textit{gold-label-aware} zero-shot prompt to get silver data from ChatGPT, which are then filtered for quality (they perform consistency verifications based on identifying specific phrases in the generated rationale). 
In contrast, we prompt \gptft (without providing the gold label, in order to obtain unbiased rationales from \gptft) to generate explanations that are structured with respect to the various linguistic features, as well as intermediate labels (which allow for easy verification of consistency); we filter the silver data by checking of consistency between the different steps of the rationale and the final label. Our method \method differs from \textsc{InstructAV} in the specific format of the output used, as well as the metrics we subsequently use to verify/ensure rationale quality - \textbf{in addition to creating accurate and high-quality models, we also have the goal of making the process of authorship verification more transparent and accessible for end-users}.  

\subsection{Authorship Analysis}
Classical AV systems use inter-textual distance to quantify the likelihood for two documents to share the same author \cite{Mosteller1984AppliedBA,Burrows2002DeltaAM,Grieve2007QuantitativeAA}. 
These verification techniques rely on expert-derived hand-crafted features to define the distance metric.
Since such stylometric methods are grounded in hand-written rules, they are inherently explainable. 
However, these methods are known to not scale well, especially in the scenario when for a given query document the correct target author needs to be mined from a large collection of candidate documents\footnote{See Table 1 of \cite{rivera-soto-etal-2021-learning} for a comparison between scores from neural transformer/convolutional models and classical \textit{tf-idf} technique.}. 

Different from classical stylometric techniques, current AV systems typically employ neural architectures which perform better at scale \cite{rivera-soto-etal-2021-learning,wegmann-etal-2022-author}.
Although significant amount of work has been devoted to advancing neural AV architectures, most of the current techniques lack interpretablity. 
For example, a typical neural AV architecture comprises of a Siamese network \cite{Koch2015SiameseNN}, where each document is separately encoded into a vector and the model is trained to ensure that the document vectors are close if the they are written by the same author and far otherwise \cite{rivera-soto-etal-2021-learning,wegmann-etal-2022-author}.
In such a network, what each dimension of the document vector \textit{means} is completely opaque. 

To bridge this dichotomy between usability and interpretibilty between classical stylometric techniques and neural AV systems, \cite{boenninghoff2019explainable} proposed to use attention heat-maps. 
This technique enabled them to determine which tokens in the input text are attended to (and not attended to) by the neural model when performing AV.
Different from attention-heat maps, model-agnostic methods, such as LIME \cite{ribeiro-etal-2016-trust}, have also been used to probe feature importance \cite{Sari2018TopicOS}.
Such explanations are helpful, however, they are inherently local\footnote{For example, heat-maps can only capture features at the token/span level.} and thus cannot capture higher level abstractions which can be informative of authorship, for example \textit{fluency}, \textit{tone and mood}, etc.

Motivated by the recent advancements made possible using LLMs, \cite{promptav} and \cite{lip} investigated whether LLMs can perform the task of authorship analysis.
They found that, when prompted correctly, the LLMs can reliably determine whether or not two documents are written by the same author. 
LLMs have also been used to construct interpretable authorship style vectors \cite{patel2023learning} and subsequently used to rewrite text from a source author in the writing style of a target author \cite{Patel2022LowResourceAS}. Similarly, LLMs have also been used by works \cite{hu2024bayesian} to perform authorship \textit{attribution}, where, given a text, the task is to identify \textit{who} wrote the text (i.e., which human or LLM from a predetermined set of authors \cite{huang2024authorship}).

Our work is the first attempt to distill free-text explanations for authorship verification from LLMs like \gptft into smaller LMs; we provide explicit focus to the joint explainability and security of \av systems via our model \method.

\subsection{Self-rationalization}
Explainability for neural network based models come in two formats: extractive rationales (includes pre-LLM era) and free-text rationales. Extractive rationales \citep{li2016understanding, sundararajan2017axiomatic, lundberg2017unified, jin2019towards} involve analyzing the influence of input tokens on the predicted output via various methods such as gradient-based analysis of input tokens ~\citep{sundararajan2017axiomatic, lundberg2017unified}, input perturbation \citep{ poerner2018evaluating, kadar2017representation}, attention heatmap analysis ~\citep{pruthi2020evaluating, stacey2022supervising, wiegreffe2019attention}, and trained models for this purpose
\citep{lei-etal-2016-rationalizing, chan2022unirex, jain2020learning, situ2021learning, liu2023decoupled}. However, extractive rationales have limited applicability as discussed previously; hence we focus on free-text rationales. 

The initial set of approaches for free-text rationales used gold standard human-written rationales to train rationale generation models \citep{camburu2018esnli,narang2020wt5, wiegreffe-etal-2021-measuring}. Following the advent of emergent self-rationalization in LLMs \cite{wei2022chain}, the research area moved to prompt large LMs with the help of curated templates with or without rationale demonstrations \citep{wei2022chain, kojima2023large, li2023making,jung2022maieutic,lightman2023lets}.
Other approaches include leverage few-shot training approaches with a handful of gold rationales \citep{marasovic-etal-2022-shot, chen2023zara}, or generating silver standard training data from large LMs to distill small LMs to be better at the task or better rationalizers. \citep{ramnath2023tailoring,li-etal-2023-symbolic, chan2023knife, wang-etal-2023-scott, saha2023language, hsieh-etal-2023-distilling}.

\section{\promptav and \chot} \label{app:cot-promptav}
\begin{table*}[h!]
    \centering
    \small
    \begin{tabular}{p{0.95\textwidth}}
    \toprule
    \textbf{\chot:} \\
    Task: On a scale of 0 to 1, with 0 indicating low confidence and 1 indicating high confidence, please provide a general assessment of the likelihood that Text 1 and Text 2 were written by the same author. Your answer should reflect a moderate level of strictness in scoring. \hl{Let's think step by step.} \\\\
    Text1: [T1] \\\\
    Text2: [T2] \\
    \bottomrule
    \end{tabular}
    \caption{\chot prompt for the task of \av, taken from \citet{promptav}}
    \label{tab:cot-prompt}
    
    \begin{tabular}{p{0.95\textwidth}}
    \toprule
    \textbf{\promptav:} \\
    Task: On a scale of 0 to 1, with 0 indicating low confidence and 1 indicating high confidence, please provide a general assessment of the likelihood that Text 1 and Text 2 were written by the same author. Your answer should reflect a moderate level of strictness in scoring. \hl{Here are some relevant variables to this problem. \\
    1. punctuation style (e.g. hyphen, brackets, colon, comma, parenthesis, quotation mark) \\
    2. special characters style, capitalization style (e.g. Continuous capitalization, capitalizing certain words) \\
    3. acronyms and abbreviations(e.g. Usage of acronyms such as OMG, Abbreviations without punctuation marks such as Mr Rochester vs. Mr. Rochester,Unusual abbreviations such as def vs. definitely) \\
    4. writing style \\
    5. expressions and Idioms \\
    6. tone and mood \\
    7. sentence structure \\
    8. any other relevant aspect \\
    First step: Understand the problem, extracting relevant variables and devise a plan to solve the problem. Then, carry out the plan and solve the problem step by step. Finally, show the confidence score}\\\\
    Text1: [T1] \\\\
    Text2: [T2] \\
    \bottomrule
    \end{tabular}
    \caption{\promptav prompt for the task of \av, taken from \citet{promptav}}
    \label{tab:promptav}

\end{table*}

Tables \ref{tab:cot-prompt} and \ref{tab:promptav} show the prompts used to apply \chot and \promptav respectively. The prompts are taken from \citet{promptav}; further note that the prompt is uniform across datasets.

\begin{table*}[ht]
    \centering
    \small
    \begin{tabular}{p{0.9\textwidth}}
    \toprule
    \textbf{Sample \chot explanation that is inconsistent with label:} \\
    \midrule
    Based on the provided texts, I assess the likelihood that Text 1 and Text 2 were written by the same author as \hlc[red!15]{0.7} (moderate confidence). \\\\
    Here's my reasoning:\\\\
1. Both texts exhibit a personal and introspective tone, which suggests that the author is comfortable sharing their thoughts and feelings.\\
2. The language and sentence structure in both texts are similar, with a mix of short and long sentences, and a conversational tone.\\
3. Both texts touch on themes of self-reflection, personal growth, and the struggle to overcome negative thoughts and emotions.\\
4. The author's writing style is informal and conversational, with a focus on expressing their thoughts and feelings rather than presenting a polished or formal argument.\\\\
However, there are some differences that suggest the texts may not have been written by the same author:\\\\
1. The tone of Text 1 is more lighthearted and humorous, while Text 2 is more serious and introspective.\\
2. The language and vocabulary used in Text 2 are more formal and philosophical, with words like `suffering,' `unrealistic standards,' and `setbacks.'\\
3. The structure and organization of the texts are different, with Text 1 consisting of short, disconnected paragraphs, and Text 2 featuring longer, more cohesive paragraphs.\\\\
Overall, while there are some similarities between the two texts, the differences in tone, language, and structure suggest that they may have been written by \hlc[red!15]{different authors}. \\
    \bottomrule
    \end{tabular}
    \caption{Sample output by \chot where the predicted label and the predicted confidence score are inconsistent with each other. The model predicts a high confidence score of 0.7 for same authorship (by usual standards of 0.5 as the threshold), but it predicts that the final answer is different authorship.}
    \label{tab:bad-cot}
\end{table*}

\begin{table*}[ht]
    \centering
    \small
    \begin{tabular}{p{0.9\textwidth}}
    \toprule
    \textbf{Sample \promptav explanation that is inconsistent with label:} \\
    \midrule
To assess the likelihood that Text 1 and Text 2 were written by the same author, we will analyze the texts based on the provided variables:\\
1. \textbf{Punctuation Style}: \\
   - Text 1 uses a variety of punctuation, including colons, commas, and quotation marks, with a distinctive style of spacing before and after punctuation marks (e.g., `this : THIS'). It also uses apostrophes in plural forms, which is unconventional (e.g., `one's').\\
   - Text 2 uses standard punctuation without the unconventional spacing or misuse of apostrophes seen in Text 1.\\
2. \textbf{Special Characters and Capitalization Style}:
   - Text 1 features continuous capitalization for emphasis (e.g., `THIS IS BETTER THAN THE ORIGINAL') and capitalizes certain words for emphasis.\\
   - Text 2 does not exhibit continuous capitalization or the same pattern of capitalizing certain words for emphasis.\\
3.\textbf{ Acronyms and Abbreviations}:\\
   - Text 1 does not prominently feature acronyms or abbreviations.\\
   - Text 2 also does not prominently feature acronyms or abbreviations, making this variable less useful for differentiation.\\
4. \textbf{Writing Style}:\\
   - Text 1 has a detailed, narrative style, sharing personal experiences and opinions in depth.\\
   - Text 2 has a more straightforward, opinion-based style without the narrative depth or personal anecdotes found in Text 1.\\
5. \textbf{Expressions and Idioms}:\\
   - Text 1 uses expressions and a more vivid language (e.g., `take 15 showers afterwards').\\
   - Text 2 uses simpler language and fewer idiomatic expressions.\\
6. \textbf{Tone and Mood}:\\
   - Text 1 has a passionate, intense tone, especially when discussing the movie's impact and the author's feelings.\\
   - Text 2 has a lighter, more casual tone, even when expressing disappointment.\\
7.\textbf{ Sentence Structure}:\\
   - Text 1 features complex sentences with multiple clauses and detailed descriptions.\\
   - Text 2 uses simpler, more straightforward sentence structures.\\
8.\textbf{ Other Aspects}:\\
   - The subject matter and interest areas differ significantly between the two texts, with Text 1 focusing on horror films and Text 2 on a Disney film. This could indicate different interests or purposes for writing, though it's not definitive evidence of different authorship.\\\\
Given these observations, the differences in punctuation style, capitalization, writing style, tone, and sentence structure suggest that Text 1 and Text 2 were \hlc[red!15]{likely written by different authors}. However, without more direct evidence (e.g., consistent unique spelling errors or very distinctive stylistic quirks that appear in both texts), absolute certainty is not possible. \\
\hlc[red!15]{\textbf{Confidence Score}: 0.8}\\
This score reflects a \hlc[red!15]{high level of confidence that the texts were written by different authors}, based on the analysis of the provided variables, but it leaves room for the possibility of the same author adapting their style to different subjects or formats." \\
    \bottomrule
    \end{tabular}
    \caption{Sample output by \promptav where the predicted label in the text and the predicted confidence score are inconsistent with each other. The model predicts a high confidence score of 0.8 (by usual standards of 0.5 as the threshold), but it predicts that the final answer is different authorship. Unless a human (or a trained model) can manually analyze the text, it is impossible to determine that the predicted label via the confidence score and the actual predicted label are different.}
    \label{tab:bad-promptav}
\end{table*}

\promptav (and by extension, \chot) which serve as the inspiration for \method produce outputs which cannot be automatically checked for consistency; these prompts produces a free-text explanation followed by a final score which can be thresholded to produce a YES/NO \av label. As observed in the experiments of \cite{promptav}, models tend to produce low confidence scores with \promptav, which means that for better accuracy, the optimal threshold in their experiments with \textsc{GPT-3.5-Turbo} was found to be 0.2-0.3, which will most probably not be consistent with the free text rationale (that is, if the rationale says that it thinks it is not confident that the authors are the same, and the confidence is 0.25, the automatic label prediction will still assume that the predicted label is YES). Further, with just the score as the label, it is still possible that we misconstrue the label because the confidence score and the model's predicted label in the text are different, as shown in the examples in Tables \ref{tab:bad-cot} and \ref{tab:bad-promptav}.

\section{\gptft silver data analysis} \label{app:silverdata}
In Table \ref{tab:silverdata} we show train dataset sizes at every stage of data generation and preprocessing. 

Further, we also perform manual analysis of \gptft's silver rationales. For the most part, \gptft generates high-quality rationales; the most frequent issue observed is \textit{hallucination} of details, especially punctuation marks. Out of the 60 samples analyzed (20 from each dataset), 10 samples showed hallucination in their rationale for punctuation style, 16 samples showed factual incorrectness across varying linguistic features, and 6 samples showed inconsistency of rationale with intermediate label in one linguistic feature (varying across samples) each.

\begin{table*}[h]
    \centering
    \begin{tabular}{|c|c|c|c|c|}
        \hline
        Dataset & Initial size & Filter for Structure & Filter for float score & Filter for \clr \\ \hline
        \imdb & 3400 & 3396 & 3255 & 2354 \\ 
        \blog & 3600 & 3600 & 3146 & 2724 \\
        \ff & 3600 & 3383 & 2887 & 2159 \\ \hline
    \end{tabular}
    \caption{Train dataset sizes at every stage of data generation and preprocessing. The `Initial size' column shows the number of datapoints generated from \gptft, and each succeeding column shows the number of datapoints remaining after the different filtering methods (in order of application).}
    \label{tab:silverdata}
\end{table*}

\section{Hyperparameters} \label{app:hparams}
We download the pre-trained \llamath from HuggingFace (\url{https://huggingface.co/meta-llama/Meta-Llama-3-8B}) and train it for all our experiments. We train it with Low-Rank Adaption (LoRA; \citealp{Hu2021LoRALA}), a parameter-efficient training method which offers lightweight training no additional inference latency, unlike other existing parameter-efficient methods \cite{Houlsby2019ParameterEfficientTL}. We train \llamath for $10$ epochs. We use no validation set, and instead use the last checkpoint obtained from training; since we have no gold-standard explanations, we opt against using the \gptft generated data as the validation set. We experiment with different values of `r' and `alpha' for LoRA (refer to \url{https://huggingface.co/docs/peft/main/en/conceptual_guides/lora} for their definitions). We use r=128 and alpha=256. Note that all our training experiments for \llamath take $\sim$12 hours, and all our test set inferences take $\sim$7-8 hours on a single NVIDIA Quadro RTX 8000 GPU. We report single-run results for all our experiments.

\textbf{Why not \llamai for fine-tuning?} As explained in the documentation for \llamai in huggingface \url{https://huggingface.co/meta-llama/Meta-Llama-3-8B-Instruct}, ``Instruction tuned models are intended for  assistant-like chat, whereas pretrained models can be adapted for a variety of natural language generation tasks''. Hence, we provide zero-shot and few-shot baselines with \llamai whereas we provide fine-tuned results on \llamath. 

\textbf{Document truncation.} When we truncate documents for GPU memory considerations in the training of \method, we always ensure that equal amounts of both documents are present in the input. Further, we always use the first chunk of the document as input and not a random chunk. 

For our baseline SVM, we use TF-IDF feature vectors for \blog (words) and \ff (characters), and word-count features for \imdb (characters). All 3 datasets' models use the RBF kernel. \imdb and \ff use n-grams of minimum size 1 and maximum size 2, whereas \blog uses only n-grams os size 2. \imdb uses a C-value of 5.0. We used sklearn version 1.2.2 for this baseline. 

For our baseline of \llamath finetuned with \chot and \promptav, our final training dataset sizes are: for \imdb, \blog and \ff, the \chot sizes - 2.7k, 2.9k, 2.4k and the \promptav sizes - 2.6k, 3k, 2.6k.

\begin{table*}[h!]
    \centering
    \small
    \begin{tabular}{|p{0.96\linewidth}|}
    \hline
    \textbf{Human Annotation Instructions:} \\ \hline
    Given two documents, the task is to say whether they are written by the same author (label: YES) or not (label: NO). The model produces a set of explanations (corresponding to different aspects like punctuation, writing style, etc.) for the same each with an intermediate YES/NO/MAYBE label, and finally a YES/NO output label for the task. \\
    The human evaluation looks at the three properties described below, and gives a score of 1, 0.5, 0 or -1 for each of the individual explanations. The "Sample human eval" section gives an example evaluation. If you give a 0.5 or a 0, please write a short explanation what was wrong with the rationale!\\
    \hline
    \end{tabular} \\
    \resizebox{\linewidth}{!}{
    \begin{tabular}{|p{0.2\linewidth}|p{0.4\linewidth}|p{0.4\linewidth}|}
    \hline
        \textbf{Criteria} & \textbf{Definition:} & \textbf{Options}  \\ \hline
        consistency with details & Is the detail consistent with the documents or hallucinated? (for example, explanation mentions parantheses when it doesn't exist in input document) & 1 - all details are consistent, 0.5 - some details are consistent, some are hallucinated, 0 - all details are hallucinated, -1 - I don't know \\ \hline
        factual correctness & Are the details factually correct? (for example, it mentions serious writing style when writing style is actually humorous) & 1 - all details are factual, 0.5 - some details are factual, 0 - no details are factual, -1 - I don't know \\ \hline
        consistency with predicted label & Is the statement faithful with the YES/NO/MAYBE at the end? & 1 - yes it is faithful, 0.5 - some details are faithful, 0 - no it is not faithful, -1 - I don't know \\ \hline
        
    \end{tabular}} 
    \caption{Instructions given to annotators for the human pilot study. Sample human evaluations are in Table \ref{tab:sample_humaneval1}.}
    \label{tab:humanevalinstr}

\end{table*}

\begin{table*}
    \small
    \begin{tabular}{p{0.95\textwidth}}
    \hline
    \textbf{Sample evaluation:} \\ \hline
    Text1: One last update. I've decided to wear my khaki pants all day. I get off work at 4:30. I see no need to rush to the 5:05 train. So I'll take my time and go on the 5:35.   I just blowdried my hair and I think it's going to be okay.  I need to eat something. I've eaten my two favorite foods, of late, for breakfast: pop tarts and pears. I think it's going to be rice and peas for lunch. \\\\
    Text2: Hey guys!  I'd like to tell you all that RICE is organising an event, 'Where is the love'.  Go to  urlLink my site  for more info, or go to the  urlLink RICE site .It's going to be great! there will be around 800 people there, and this event will be about how the world's gone wrong with terrorism, war... etc.   Date: 10 July 2004 Place: Willoughby Civic Centre Time: 7pm  Ask Ben, Adro or me for more info.  We'd like  All  of you to come!   urlLink\\\\
    GOLD LABEL: NO\\\\
    \end{tabular}
    \begin{tabular}{|p{0.55\textwidth}|p{0.1\textwidth}|p{0.1\textwidth}|p{0.1\textwidth}|} 
    \hline
    \textbf{Rationale} & \textbf{P1} & \textbf{P2} & \textbf{P3} \\ \hline
    \hl{punctuation style}: Both texts use a variety of punctuation marks such as periods, commas, and ellipses. However, Text 1 uses more complex structures like colons and parentheses, which are less prevalent in Text 2. \colorbox{yellow}{\underline{MAYBE}} & 0.5 (ellipses / parantheses hallucinated) & 0.5 (colons in both) & 1 \\ \hline
    \hl{special characters style, capitalization style}:Text 1 shows a more natural capitalization style without unusual emphasis on continuous capitalization or certain words. Text 2 capitalizes 'RICE' and uses all caps for 'URL', indicating a different approach. \colorbox{yellow}{\underline{NO}} & 0.5 (url not capitalized in Text2) & 0.5 (RICE is implied as an organization, it it not a reflection of the capitalization approach) & 1 \\ \hline
    \hl{acronyms and abbreviations}: Text 1 does not prominently feature acronyms or abbreviations, while Text 2 includes 'URL' as an abbreviation. This suggests a difference in the use of abbreviations. Conclusion: \colorbox{yellow}{\underline{NO}} & 1 & 1 & 1 \\ \hline
    \hl{writing style}: Text 1 has a more personal, reflective style, discussing personal routines and thoughts. Text 2 adopts a more informative and promotional tone, inviting people to an event. The styles are quite distinct. \colorbox{yellow}{\underline{NO}} & 1 & 1 & 1 \\ \hline
    \hl{expressions and idioms}: Neither text heavily relies on idiomatic expressions, but the context in which they write differs significantly, with Text 1 focusing on personal narrative and Text 2 on event promotion. \colorbox{yellow}{\underline{NO}},  & 1 & 1 & 1 \\ \hline
    \hl{tone and mood}: The tone of Text 1 is casual and somewhat contemplative, while Text 2 is more enthusiastic and outward-facing, aiming to promote an event. These differences suggest different authors. \colorbox{yellow}{\underline{NO}} & 1 & 1 & 1 \\ \hline
    \hl{sentence structure}: Text 1 uses a mix of short and long sentences, with a more conversational flow. Text 2 also uses a mix but is more structured, likely due to its promotional nature. The sentence structures show some similarities but also differences in context and purpose. \colorbox{yellow}{\underline{MAYBE}} & 1 & 1 & 1 \\ \hline
    \hl{any other relevant aspect}: TThe subject matter of the two texts is vastly different, with Text 1 focusing on personal habits and reflections, and Text 2 on organizing an event and expressing concerns about global issues. This difference in content and focus further supports the likelihood of different authors. \colorbox{yellow}{\underline{NO}} & 1 & 0.5 (difference in content doesn't mean different authors) & 0 (should be MAYBE. Difference in content is not difference in authorship) \\
    \hline
    \end{tabular}   
    \caption{Sample human evaluation shown to human annotators}
    \label{tab:sample_humaneval1}

\end{table*}

\begin{table}[h!]
\centering
\resizebox{\linewidth}{!}{%
\begin{tabular}{lccccccccc}
\toprule
 \multirow{2}{*}{Linguistic Features} & \multicolumn{3}{c}{\imdb} & \multicolumn{3}{c}{\blog} & \multicolumn{3}{c}{\ff} \\
 \cmidrule(lr){2-4} \cmidrule(lr){5-7} \cmidrule(lr){8-10}
  & P1 & P2 & P3 & P1 & P2 & P3 & P1 & P2 & P3 \\ \midrule
punctuation style & 15 & 31  & 43 & 25 & 38 & 45 & 23  & 34 & 37 \\  \midrule
special characters/ & \multirow{2}{*}{31}  & \multirow{2}{*}{40} & \multirow{2}{*}{47} & \multirow{2}{*}{23} & \multirow{2}{*}{37} & \multirow{2}{*}{46} & \multirow{2}{*}{33} & \multirow{2}{*}{39} & \multirow{2}{*}{43} \\
capitalization style &  &  & & & & &  & &  \\  \midrule
acronyms/abbr. &  38 &  41 &  47 & 25 & 37 & 45  & 47  & 46 & 43 \\ \midrule
writing style & 47  & 45  & 46  & 44 & 48 & 49 &  45& 44 & 50 \\ \midrule
expressions/idioms & 44  &  48 & 49  &  40 & 43 & 45 & 46 & 46 & 44\\ \midrule
tone and mood & 47  &  48 &  48 & 46 & 47  & 48 & 47  & 43 & 49 \\ \midrule
sentence structure & 45  & 48  &  46 & 42 & 45 & 48 & 39 & 40 & 47 \\ \midrule
any other &  48 &  43 & 38  & 46 & 38 & 37 & 48 & 46  & 45 \\ \bottomrule
\end{tabular}%
}
\caption{\textbf{Human Pilot for individual datasets:} This table shows the number of examples (out of 20) where all 3 annotators agree that the rationale (corresponding to the linguistic feature) conforms to the property being analyzed. P1 is `Detail-Consistency', P2 is `Factual-Correctness' and P3 is `Label-Consistency'.}
\label{tab:humaneval}
\end{table}

\section{\method with different offline models, Practical advantages of \method} \label{app:other-local-models}
\begin{table*}[h!]
\centering
\resizebox{0.9\textwidth}{!}{%
\begin{tabular}{ccccccccc}
\toprule
\multirow{2}{*}{Model} & Test $\rightarrow$  & \multicolumn{2}{c}{\imdb} & \multicolumn{2}{c}{\blog} & \multicolumn{2}{c}{\ff} \\
\cmidrule(lr){2-2} \cmidrule(lr){3-4} \cmidrule(lr){5-6} \cmidrule(lr){7-8}
 & Train $\downarrow$ & \accs & \conss & \accs & \conss & \accs & \conss \\ \midrule
   \llamath  & Fine-tune & \cellcolor{lviolet}\textbf{74.1} & \cellcolor{lgreen}\textbf{0.99}$^*$ & \cellcolor{lviolet}\textbf{60.6} & \cellcolor{lgreen}\textbf{1.00}$^*$ & \cellcolor{lviolet}\textbf{62.6}$^*$ & \cellcolor{lgreen}\textbf{0.99}$^*$ \\ \midrule
   \textsc{Mistral-7B} & Fine-tune & 73.7 & 0.99 & 59.2 & 0.99 & 57.7 & 0.98 \\ \midrule
   Stable-LM-Zephyr-3B & Fine-tune & 50.8 & 0.71 & 47.7 & 0.81 & 42.0 & 0.68 \\ 
   \bottomrule
\end{tabular}
}
\caption{Empirical Results comparing \method trained with local models of different sizes. \textsc{Mistral-7B} is the next best performing model after \llamath. \textsc{Stable-LM-Zephyr-3B} however has a lower performance.}
\label{tab:localmodelexps}
\end{table*}
We report experiments with local models of varying sizes in Table \ref{tab:localmodelexps}. After \llamath, \textsc{Mistral-7B} is the next best performing model. 

\paragraph{Practical Advantages:} \method also demonstrates practical advantages over \gptft such as (1) Energy: \llamath is just 8B parameters, whereas \gptft is orders of magnitude bigger - \method hence consumes much less energy per datapoint, and hence is more affordable and efficient, (2) Financial cost: \method is also more financially affordable since it can be hosted on a local server; \gptft or any API based language model would cost much more since it charges for every input/output token, (3) Speed: On average, \method takes 3 minutes per datapoint for inference, whereas \gptft takes 4.5 minutes.
\subsection{Human Pilot} \label{sec:humaneval}
Prior works in this field largely focus on task accuracy, with much lesser focus on explainability \cite{patel2023learning} and by extension, human study of the explanations. In this work, we take the \textit{first step towards human analysis of \av explanations} by providing a pilot study on \method's rationales.


\section{Human Evaluation} \label{app:humaneval}
Table \ref{tab:humanevalinstr} shows the instructions given to the annotators regarding the properties to be analyzed and Table \ref{tab:sample_humaneval1} shows a sample human evaluation. We perform an in-house human evaluation, and annotators used were proficient in English and well aware of the high complexity of the task. 

Table \ref{tab:humaneval} shows the human pilot study numbers for individual datasets. The aggregate of these numbers were used for the overall analysis in Appendix \ref{sec:humaneval}.

\section{Random Errors due to Computational Limitations} \label{app:randomerrors}
We acknowledge random errors (i.e., not systematic) that could have occurred in our results owing to limitations of language models and computational resources. Our method uses a language model to read two input documents and provide an output; however, in practice, language models have a fixed maximum length of input due to model limitations as well as GPU memory limitations. This means that if the two input documents together (along with the instructions) do not fit into the input, it becomes necessary to truncate them, which leads to loss of information of their latter parts. This issue is exacerbated when we move towards using in-house models on user servers (which typically have a shorter context window than API-based models like \gptft). In our experiments with \llamath, we used just 2 GPUs (to have a realistic user environment), which means that while training we had to restrict the input to be of maximum length $1280$\footnote{note: during inference, we did not limit the size of the input, but this could pose a problem if the documents became much longer}, which could have affected the training for the longer documents. Further, even for generating the training data from \gptft for \ff, the excessive size of the documents made the \gptft generations economically infeasible; we cut our the input to to a maximum of 300 words per document, which could have created errors/incomplete rationales that eventually propagated to our downstream training of \llamath.  

\section{Out-of-domain inference} \label{app:ood_inf}

\begin{table}[h!]
    \centering
    \resizebox{\linewidth}{!}{
    \begin{tabular}{ccccccc}
        \toprule
        Test $\rightarrow$  & \multicolumn{2}{c}{\imdb} & \multicolumn{2}{c}{\blog} & \multicolumn{2}{c}{\ff} \\
        \cmidrule(lr){1-1} \cmidrule(lr){2-3} \cmidrule(lr){4-5} \cmidrule(lr){6-7}
        Train $\downarrow$ & \accs & \conss & \accs & \conss & \accs & \conss \\ \midrule
        \imdb & --- & --- & 59.9 & 1.0 & 61.2 & 0.99 \\ 
        \blog & 71.3 & 0.96 & --- & --- & 45.3 & 0.77 \\
        \ff & 68.8 & 1.0 & 59.3 & 1.0 & --- & --- \\
        \bottomrule
    \end{tabular}}
    \caption{Out of domain performance by \method}
    \label{tab:ood}
\end{table}
\noindent Finally, we evaluate \method models on the other datasets' test sets (refer Table \ref{tab:ood}); we notice robust out-of-domain performance especially by the \imdb and \ff models. This is a significant result, as it shows the generalizability of this approach to unseen \av datasets: although we use fine-tuning to improve performance over a zero-shot approach, it is not fine-tuning specifically to the domain that is most critical but rather fine-tuning to the structured format of \method's output.

\section{Post-hoc Interpretability}
We finally clarify that \method was designed to serve \textit{post-hoc rationalization}, and not mechanistic (i.e., low-level mathematical) understanding of the language model's workings. We quote from \citet{lipton2018mythos} to emphasize that \method's improved post-hoc explainability leads to improved utility to (1) downstream users, and (2) researchers, who can leverage \method's accessible outputs for detailed analyses, to address errors and develop improved models: ``\textit{while post-hoc explanations often do not elucidate precisely how a model works, they may nonetheless confer useful information for practitioners and end users of machine learning}''.

\section{More rationale properties} \label{app:more-rat}
In addition to consistency, there are several properties necessary for high quality rationales \cite{joshi-etal-2023-machine,ramnath2023tailoring}. Some of these are easily verifiable by humans, even if there exist no prevalent automatic methods to measure them: for eg:- factual correctness, hallucination in the rationale, etc. that we measured in our human evaluation (\textsection\ref{sec:humaneval}). There are also properties that are hard to verify even by human evaluators: for eg:- \textit{completeness} of the rationale, i.e., has the rationale missed any significant (dis)similarity between the documents?. In fact, completeness does not have a comprehensive definition even for simpler tasks like multi-choice QA \cite{ramnath2023tailoring}. In the future, we would like to work on automatic metrics and model training for these properties specifically for \av. 

\section{Analyzing cases where \promptav succeeds but \method fails}
We consider the $88$ samples of the \blog test set where \promptav succeeds but \method fails. From qualitative analysis of a few samples, we find out the following (non-exhaustive) patterns: 
\begin{itemize}
    \item Cases where the gold label is `YES', and both methods fail: \method wrongly predicts a label of `NO' with a final score of $<0.5$. \promptav gives the wrong reasoning, but concludes by predicting a confidence score $>0.5$ that the authors are \textit{different} which leads to a misinterpretation that \promptav was right. We provide an example of this error by \promptav in Appendix \ref{app:cot-promptav}.
    \item Differences in perceiving patterns such as idioms, acronyms: Both methods generate similar explanations for abstract features such as writing style, tone \& mood. But when it comes to token-level features such as acronyms, idioms, etc., the methods sometimes predict different subsets, which leads to differing final predictions. This comes under the issue of \textit{completeness} of rationale, which we discuss in the Limitations section and in Appendix \ref{app:more-rat}. In this particulat situation, both models have provided different incomplete rationales, which eventually leads to different labels.
\end{itemize}

\end{document}